\documentclass{vgtc}
\usepackage{amsmath}
\usepackage{setspace}
\usepackage{titlesec}
\usepackage[percent]{overpic}

\newcommand{\CHANGED}[1]{\textcolor{black}{#1}}

\graphicspath{{figures/}{pictures/}{images/}{./}} 

\usepackage{times}                     

\usepackage{tabu}                      
\usepackage{booktabs}                  
\usepackage{lipsum}                    
\usepackage{mwe}                       

\usepackage{mathptmx}                  
\usepackage{adjustbox} 
\usepackage{placeins}

\onlineid{0}

\vgtccategory{Research}

\vgtcinsertpkg

\title{Evaluating `Graphical Perception' with Multimodal LLMs }

\author{Rami Huu Nguyen\thanks{e-mail: rami@mpsych.org}\\ %
        \scriptsize University of Massachusetts Boston \and
Kenichi Maeda\thanks{e-mail: kenichi.maeda001@umb.edu}\\ %
        \scriptsize University of Massachusetts Boston %
\and Mahsa Geshvadi\thanks{e-mail: mahsa.geshvadi001@umb.edu}\\ %
     \scriptsize University of Massachusetts Boston %
\and Daniel Haehn\thanks{e-mail: daniel.haehn@umb.edu}\\ %
     {\scriptsize \centering University of Massachusetts Boston}}

\teaser{
  \centering
  \vspace*{-0.6 cm}
  \adjustbox{center=0.8\textwidth}{%
    \includegraphics[width=1.15\textwidth]{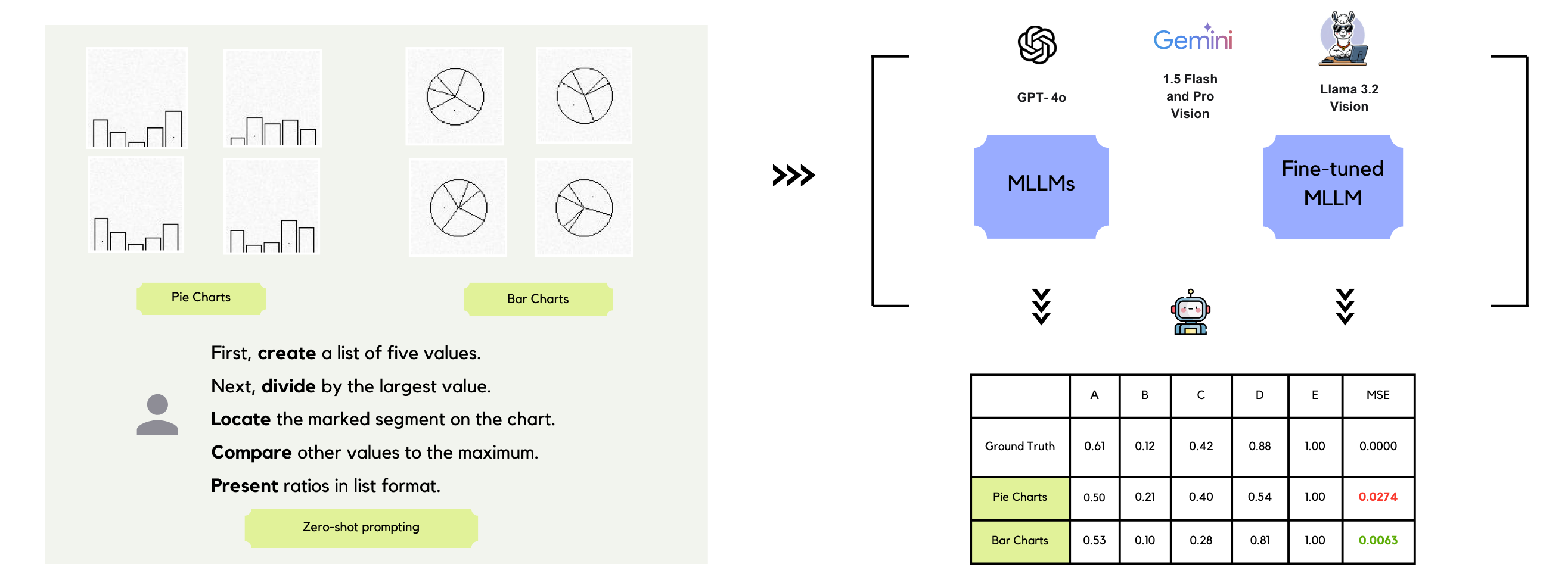}
  }
  \label{fig:teaser}
  \caption{\textbf{Computing Cleveland and McGill’s Position-Angle Experiment using Multimodal Large Language Models}. We replicate the original experiment by asking MLLMs to interpret values in pie and bar charts using zero-shot prompting, where models follow instructions without prior examples. \CHANGED{Results highlight that MLLMs predict values more accurately from bar charts (mean squared error (MSE) in green).}
}}

\abstract{

Multimodal Large Language Models (MLLMs) have remarkably progressed in analyzing and understanding images. Despite these advancements, accurately regressing values in charts remains an underexplored area for MLLMs. For visualization, \textcolor {blue} {\textbf{how do MLLMs perform when applied to graphical perception tasks?}} Our paper investigates this question by reproducing Cleveland and McGill's seminal 1984 experiment and comparing it against human task performance. Our study primarily evaluates fine-tuned and pretrained models and zero-shot prompting to determine if they closely match human graphical perception. Our findings highlight that MLLMs outperform human task performance in some cases but not in others. We highlight the results of all experiments to foster an understanding of where MLLMs succeed and fail when applied to data visualization.
    
} 


\begin{document}


\firstsection{Introduction}

\maketitle
Nowadays, \CHANGED {data visualization} has become increasingly important in our lives \cite{kafle2018dvqaunderstandingdatavisualizations, meng2024chartassisstantuniversalchartmultimodal}. There has been a rising research focus on computational techniques for studying \CHANGED{charts, and graphs.}  \cite {haehn2018evaluating,kafle2018dvqaunderstandingdatavisualizations, kahou2018figureqaannotatedfiguredataset}, which are applied in several applications, including data extraction, classification, visual Q\&A (e.g., “computer, which section is greater?”), and design evaluation or synthesis. MLLMs have made significant progress in analyzing and understanding images \cite{agrawal2016vqavisualquestionanswering, guo2024understandinggraphicalperceptiondata,han2023chartllamamultimodalllmchart,luo2024mmevolempoweringmultimodallarge,lv2024kosmos25multimodalliteratemodel,wang2024m2ptmultimodalprompttuning, zeng2024advancingmultimodallargelanguage}. Although MLLMs perform well in understanding charts, they struggle in generalization and face difficulties accurately answering chart-related questions \cite{han2023chartllamamultimodalllmchart,meng2024chartassisstantuniversalchartmultimodal}. This requires the MLLMs to understand both language and information derived in charts and apply reasoning skills to provide correct answers \cite{du2023makesgoodvisualinstructions,liu2024mmc}. 
\noindent Most current MLLMs are pre-trained vision and knowledge, which means those models are trained before with general knowledge, and they might struggle with new application \cite{han2024multimodallargelanguagemodels,lee2024multimodalreasoningmultimodalknowledge,liu2023visualinstructiontuning,wang2023visionllmlargelanguagemodel}, which potentially lead to incorrect visual understanding. Understanding images (computer vision) poses unique challenges as compared to understanding language \cite{wang2023visionllmlargelanguagemodel} \cite{xie2023agicomputervisionlessons}. Language often relies on structured syntax and grammar, while chart data depends on spatial relationships, patterns, and context \cite{wang2024pictureworththousandwords}. Hence, analyzing chart data might be more challenging for the MLLMs. 
\noindent What's more, the limitation of MLLMs also persists: MLLMs find it difficult to recognize small objects or tiny details in pictures \cite {bai2023qwenvlversatilevisionlanguagemodel,guo2024understandinggraphicalperceptiondata,zhang2024exploringperceptuallimitationmultimodal}. Additionally, MLLMs currently have difficulty pinpointing the important details in the images that are unclear or absent in the images \cite{wu2023vguidedvisualsearch}. Also, humans use senses such as sight and language to understand the world and recognize new objects based on their knowledge.\cite{liu2023visualinstructiontuning, wang2024m2ptmultimodalprompttuning}. Zero-shot prompting follows similar principles as human abilities, with its main purpose being to improve MLLMs using the zero-shot prompts to make them perform better without the need for additional training. Cleveland and McGill introduced the concept of graphical perception, explaining how humans visually interpret information from graphs \cite{Cleveland_84,cleveland1985graphical}. \CHANGED {Cleveland and McGill defined elementary perceptual tasks as mental-visual processes and ranked how complex those tasks are for humans. Building on their work, our research is to compare fine-tuned MLLMs, trained for specific low-level graphical perception tasks, with pretrained MLLMs models using zero-shot prompting. Once those MLLMs models perform well on those tasks, they will establish a strong foundation for interpreting more complex visualizations.}

\subsection{Related Work}

To study this graphical perception concept, Cleveland and McGill performed the position-angle experiment (comparing pie charts and bar charts) and the position-length experiment (where participants were asked to compare values in groups and divided charts \cite{Cleveland_84,cleveland1985graphical}.  Then, the authors use this to redesign a statistical map via bars, framed rectangles, and Weber’s law \cite{harrison2014ranking}, using the proportional relation between an initial distribution density and perceivable change. Heer and Bostock later reproduced Cleveland and McGill’s experiments by crowdsourcing participants on Amazon Mechanical Turk with similar findings \cite{heer2010crowdsourcing}. Harrison et al. also repeated the studies while studying viewer emotions, again with similar results \cite{harrison2014ranking}. Talbot et al. explored how variation in bar charts affects human prediction \cite{Talbot2014FourEO}. Cleveland and McGill’s idea of graphical perception does not rely on human-specific traits, and it targets the process of decoding information visually. Thus, it allows machines to perform similar tasks, such as MLLMs. Nonetheless, machines must match humans' graphical perception levels to function effectively in practice. Our research paper is inspired by Cleveland and McGill's, and Haehn et al. \cite {haehn2018evaluating} also use CNNs for similar stimuli to investigate where machines perceive and reason visual relationships similar to human perception. 

\section{\textbf{Experiment Setup}}

\CHANGED{We compare MLLMs to human baselines across five experiments}. \textbf{E1} estimates quantities from visual features based on Cleveland and McGill’s elementary perceptual tasks. \textbf{E2} replicates their position-angle experiment, comparing pie and bar charts. \textbf{E3} reproduces their position-length experiment, analyzing grouped versus divided bar charts. \textbf{E4} evaluates bars and framed rectangles using their visual cue framework. \textbf{E5} conducts a Weber’s law point cloud experiment.

\subsection{\textbf{Networks and Processes}}

\CHANGED{As a starting point, we used three latest closed-source pretrained MLLMs, including \textbf{GPT-4o} with 1.8 trillion parameters, \textbf{Gemini 1.5 Flash} with 8 billion parameters, \textbf{Gemini 1.0 Vision Pro} (unavailable parameter data), and one open-source \textbf{Llama 3.2-Vision} with 11 billion parameters.} For \textit{fine-tuned MLLMs}, we used \textbf{Llama 3.2-Vision}, with 6.0 billion parameters, of which 94.4 million were trainable. We also produced 15 fine-tuned MLLMs (each experiment has three fine-tuned MLLMs) for this study. View our fine-tuning details in our supplementary  material. Each experiment generated \textit{5000 unique training images}, \textit{1000 for unique validation}, and \textit{55 for unique testing for each task}, add \textit{added 5\% noise} to each image, and ensure no data leakage. Detailed dataset generation and preprocessing are discussed in our supplementary material.

\subsection{\textbf{Measurements}}

Our research calculates models’ performances using the midmean logistic absolute error metric (MLAE). Our research paper is inspired by Cleveland and McGill's methodology in their 1984 study \cite {Cleveland_84} and defines our calculation as follow:

\[{
\text{MLAE} = \frac{1}{N} \sum_{i=1}^{N} \log_2 \left( \lvert \text{predicted}_i - \text{true }_i \rvert + 0.125 \right)
}\]

\noindent We also include standard error metrics such as mean squared error (MSE) and mean absolute error (MAE). Our fine-tuned models use Cross-Entropy loss instead of MLAE, applying the logarithm before averaging to ensure fair evaluation of small and large errors.  d

\subsection{\textbf{Stimuli}}

We employed the stimuli generator designed by Haehn et al. \cite{haehn2018evaluating} for each perceptual task, and the number of parameter values varied depending on each experiment.

\subsection{\textbf{Human Baselines}}

We gather human baseline measurements for the position-angle (E2) and position-length (E3) experiments from \cite{Cleveland_84}, with 51 participants. We also included human baseline measurements from Heer and Bostock’s study \cite{heer2010crowdsourcing}, with 50 participants. In both experiments, each participant reviewed ten stimuli under each condition. We followed Haehn et al.'s paper \cite{haehn2018evaluating} for E1, E4, and E5. \CHANGED {The human baseline in Haehn et al.'s paper were gathered from 25 participants on Amazon Mechanical Turk for those three experiments. Each experiment has ten stimuli for participants (nine for E1, two for E4, and three for E5), including three practice stimuli for each condition. }

\subsection{\textbf{Data Preprocessing}}

Initially, we collected \textbf{825 responses per task across all five experiments}, with each experiment has three runs. However,  \textit{invalid and missing responses} were found, primarily for pretrained models from E1 to E5 in each experiment. Therefore, we removed those invalid values from each initial dataset. To ensure fair comparisons between models, we excluded these invalid responses. However, since each model produced a different number of invalid responses, this resulted in an imbalance total number of valid responses per model.  Since these invalid responses vary in each run, we balanced the dataset by randomly choosing the global minimum number of valid responses. Examples of our invalid responses and how we balanced the dataset are provided in our supplementary al material.

\section{Experiment 1}

Cleveland and McGill introduced ten elementary perceptual tasks that use graphical elements or visual marks to represent numerical values \cite{Cleveland_84}. These tasks are the low-level building blocks for information visualizations. Examples are \textbf{estimating position on a common scale, position on non-aligned scales, length, direction (or slope), angle, area, volume, curvature, and shading (or ink density)}. To evaluate these tasks, we developed a visual representation as 100 x 100 raster images. Our main goal is to evaluate whether our MLLMs networks can regress the numerical value they encoded. We also added complexity to each task by generating multiple versions of elementary perceptual tasks. For instance, we created multiple angle degrees by changing the line's direction and the angle's size. 

\begin{figure} [h!]
    \centering
    \hspace*{1 cm}
    \vspace{-0.3 cm}
    \scalebox{1}{
        \includegraphics[width=\linewidth]{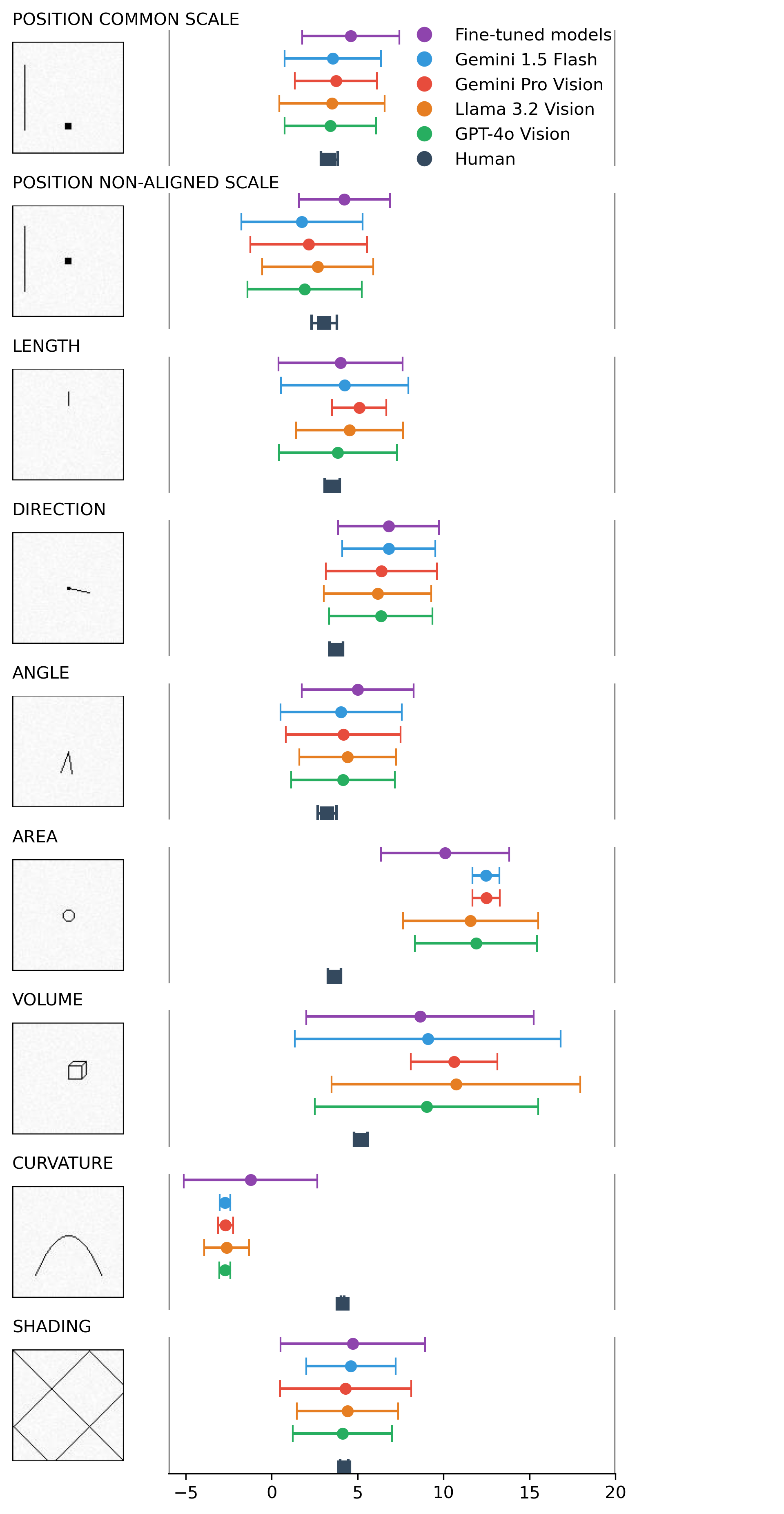} 
    }
    \caption{\textbf{Elementary perceptual tasks results for the most complex task parameterization}. 
    In each column: Left: Example stimuli image. Right: MLAE and bootstrapped 95\% confidence intervals for different networks. Lower MLAE scores are better.}
    \label{fig:enter-label}
    \FloatBarrier
\end{figure}

\subsection{\textbf{Hypothesis and Results}}

\textbf{H1.1: Our fine-tuned MLLMs model can regress quantitative variables from graphical elements and outperform pretrained models and human \CHANGED {graphical} perception on these elementary perceptual tasks. }

Figure 2 highlights our fine-tuned models have the highest MLAE errors compared to human perception in most tasks. In \textbf{\textit{angle tasks}}, our fine-tuned models achieve \textbf{MLAE = 5.01}, MAE = 51.65, SD = 1.66, while humans outperform them with \textbf{MLAE = 3.22}, SD = 0.54. In \textbf{\textit{area tasks}}, our fine-tuned models struggle with \textbf{MLAE = 10.09}, MAE = 1815.65,  SD = 1.91, while humans perform better (\textbf{MLAE = 3.64}, SD =  0.38).  In \textbf{\textit{volume tasks}}, our fine-tuned models show  \textbf{MLAE = 8.63}, MAE = 2034.04, SD =  3.38, while humans perform better (\textbf{MLAE = 5.18}, SD = 0.40). In particular, the fine-tuned models only performed the best with \textbf{\textit{\textcolor{blue}{curvature}}} task with an \textbf{\textcolor{blue}{MLAE of -1.23, MAE = 1.19, SD = 1.98}}; however, it also underperformed all pretrained MLLMs with average MLAE ranging from -2.73 to -2.62. Our fine-tuned models only performed slightly better than two or three pretrained models, especially in area and volume. For instance, in the \textbf{\textit{volume}} \textbf{tasks}, the difference is minimal as our fine-tuned MLLMs record an \textbf{MLAE of \textbf{8.63}}, slightly higher than the pretrained Gemini 1.5 Flash, Gemini Pro Vision, Llama 3.2-Vision and GPT-4o which have \textbf{MLAE scores of 9.08, 10.62, , 10.72 and 9.01}, respectively. \CHANGED{Across tasks, we compare the average regression performance of our networks and report statistically significant effects (\textbf{F = 10.303, }\textbf{$p < 0.01$}). Tukey's HSD post-hoc test highlights that \textbf{Gemini 1.5 Flash and GPT-4o significantly outperform our fine-tuned models}, while the difference between our fine-tuned models and Gemini 1.5 Flash and Llama 3.2-Vision is not statistically significant (\textbf{$p > 0.01$}).} Therefore, \textbf{we do not accept H1.1}.

\section{Experiment 2}

Cleveland and McGill compare bar and pie charts by studying how humans perceive position ratios and angles \cite{Cleveland_84}. Following Cleveland and McGill's proposed encoding, we generate rasterized images to evaluate how our networks perceive these two tasks. Each visualization consists of pie or bar charts representing numbers that sum to 100, with individual numbers ranging from 3 to 39. We modified our approach to minimize value differences. Cleveland and McGill created stimuli with a minimum scale difference of 0.1, but our models only process 100 x 100-pixel images as input; we can only minimally represent a difference of 1 pixel. In Cleveland and McGill's experiments \cite{Cleveland_84}, participants were asked to estimate the ratio of the four smaller segments to the known and marked most significant segments. Similarly, we marked the largest segment in each visualization with a single pixel dot. We tasked our networks with performing multiple regressions to estimate the ratio of the remaining four segments. 

\subsection{Hypothesis and Results}
\begin{figure}[h!]
    \hspace*{0.5 cm}
    \vspace{0.2 cm}
    \centering
    \scalebox{1.15}{%
        \includegraphics[width=1\linewidth]{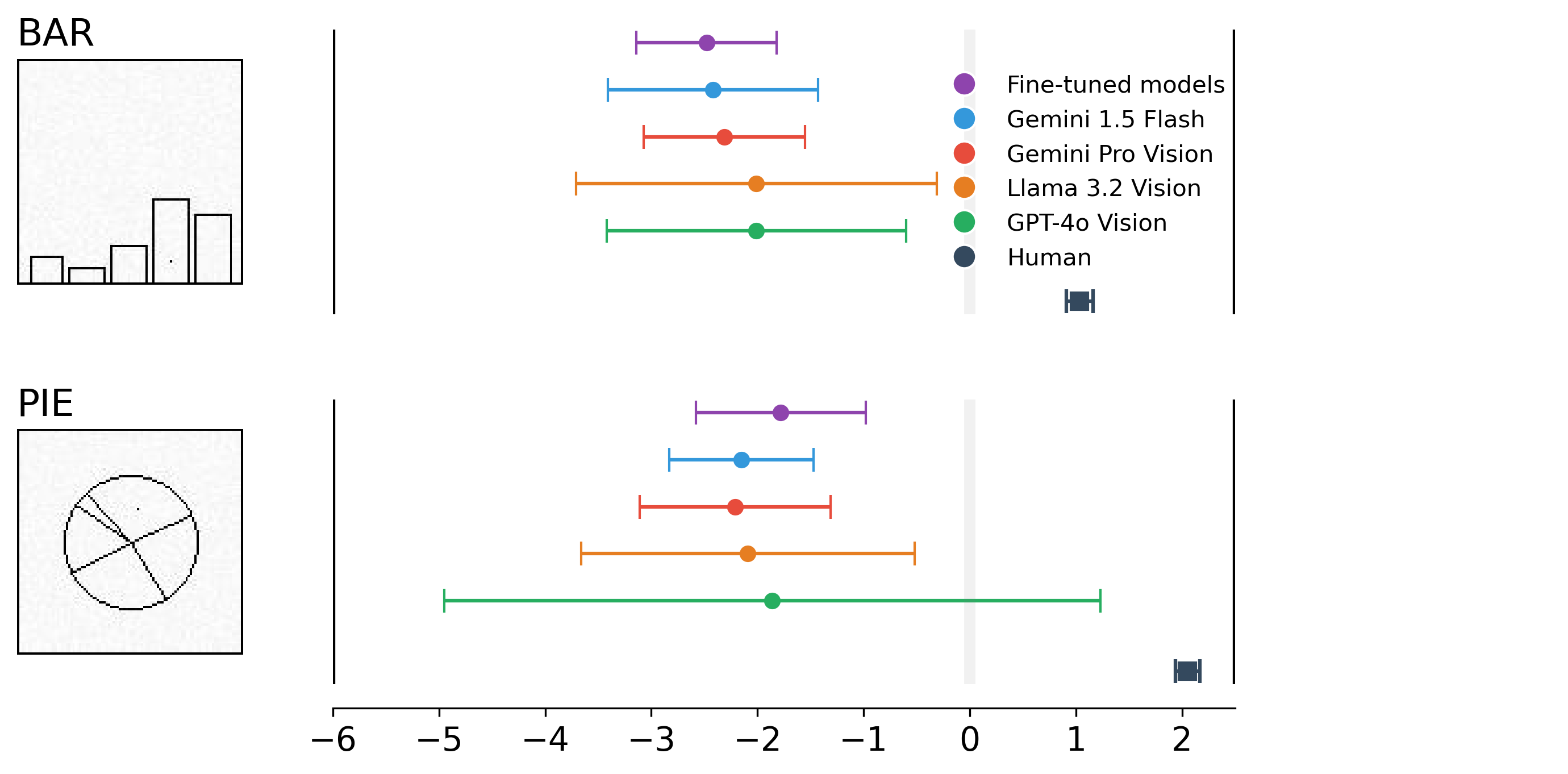}
    }
    \caption{\textbf{Computational results of the position-angle experiment}.
Left: Example stimuli. Right: MLAE and bootstrapped 95\% confidence intervals (the lower, the better)}
    \label{fig:enter-label}
\end{figure}

\noindent\textbf{H2.1: Our fine-tuned models outperform human \CHANGED {graphical} perception for bar and pie charts tasks.
}
\\\indent For the \textbf{bar chart }tasks, our \textit{fine-tuned models} achieved \textbf{an MLAE of -2.48} (SD =  0.34, MAE: 0.06). For our pretrained models, \textit{Gemini 1.5 Flash} achieved \textbf{an MLAE of -2.42} (SD =  0.51) and an MAE of 0.12, \textit{Gemini 1.0 Pro Vision} showed \textbf{an MLAE of -2.31} (SD = 0.39) and MAE of 0.08, \textit{Llama 3.2-Vision} had an \textbf{MLAE of  -2.01}, (SD = 0.87) and MAE of 0.28, and \textit{GPT-4o} had  an MLAE of  -2.01, (SD = 0.72) and MAE of 0.28. Comparing this to \textbf{the human baseline of MLAE: 1.035} (SD = 0.115) for bar charts, both pre-trained and fine-tuned models outperform human graphical perception, as a lower MLAE suggests better performance.

For the \textbf{pie chart} tasks, our \textit{fine-tuned models} achieved \textbf{an average MLAE of -1.78} (SD = 0.41) and an MAE of 0.18. For our pretrained models, \textit{Gemini 1.5 Flash} achieved \textbf{an MLAE of -2.15} (SD =  0.34) and MAE of 0.11, \textit{Gemini 1.0 Pro Vision} showed \textbf{an MLAE of -2.21} (SD =  0.46) and MAE of 0.10, Llama 3.2-Vision had an \textbf{MLAE of  -2.09}, (SD = 0.80 and MAE of 0.28, and \textit{GPT-4o} had an MLAE of  -1.86, (SD = 1.58) and MAE of 1.22. Comparing this to \textbf{the human baseline of MLAE: 2.05} (SD = 0.125) for pie charts, both pretrained and fine-tuned models outperform human graphical perception, as a lower MLAE suggests better performance. Based on the results from two tasks, \textbf{we accept H2.1}. 

\singlespacing

\textbf{H2.2: Pie charts are more challenging for pre-trained and fine-tuned models than bar charts. }

Figure 3 reveals that \textcolor{blue}{\textbf{our fine-tuned models outperformed pretrained models for bar tasks}}. Pie charts also have higher the average MLAE values than bar charts. \CHANGED {Across tasks, we compare the average regression performance of our networks and report statistically significant effects (F = 25.614, $p < 0.01$). Tukey’s HSD post-hoc test highlight that our fine-tuned models performs significantly worse than Gemini 1.5 Flash and Gemini 1.0 Pro Vision ($p < 0.01$), but our fine-tuned models only outperforms GPT-4o ($p < 0.01$) with a mean difference of 0.15. Additionally, the difference between our fine-tuned models and Llama 3.2-Vision is not statistically significant ($p > 0.01$), indicating similar performance. Since the results do not fully support the hypothesis for across models, we \textbf{partially accept H2.2}.}

\section {Experiment 3}

Cleveland and McGill evaluated the perception of position and length across five variations of the group and divided bar charts \cite {Cleveland_84}. While both charts highlight identical information, the perceptual tasks are interpreted differently. A group of bar charts always requires the judgments of positions along a common scale, while a group of divided bar charts also involves length judgments.  Types 1, 2, and 3 focus on judging positions along a common scale, whereas types 4 and 5 require length judgment. In their experiment, participants were asked to estimate the percentage of the smaller marked bar element of the larger one. Cleveland and McGill ranked the tasks from easiest (Type 1) to hardest (Type 5). We followed their method to generate data and created ten value pairs using the equation:
\[
s_i = 10 \times 10^{\frac{i-1}{12}}, \quad i = 1, \dots, 10
\]
The experiment involves a dataset comprising a combination of unique labels across its subsets. For each chart type, we created visualization charts corresponding to these ground truth values. We ask our MLLMs to follow zero-shot prompts to estimate the percentage of the smaller value relative to the larger one, treating this as a single-value regression problem.

\begin{figure} [h!]
    \centering 
    
    \hspace*{1.0 cm}
    \vspace{-0.5 cm}
    \scalebox{1.0}{%
        \includegraphics[width=1\linewidth]{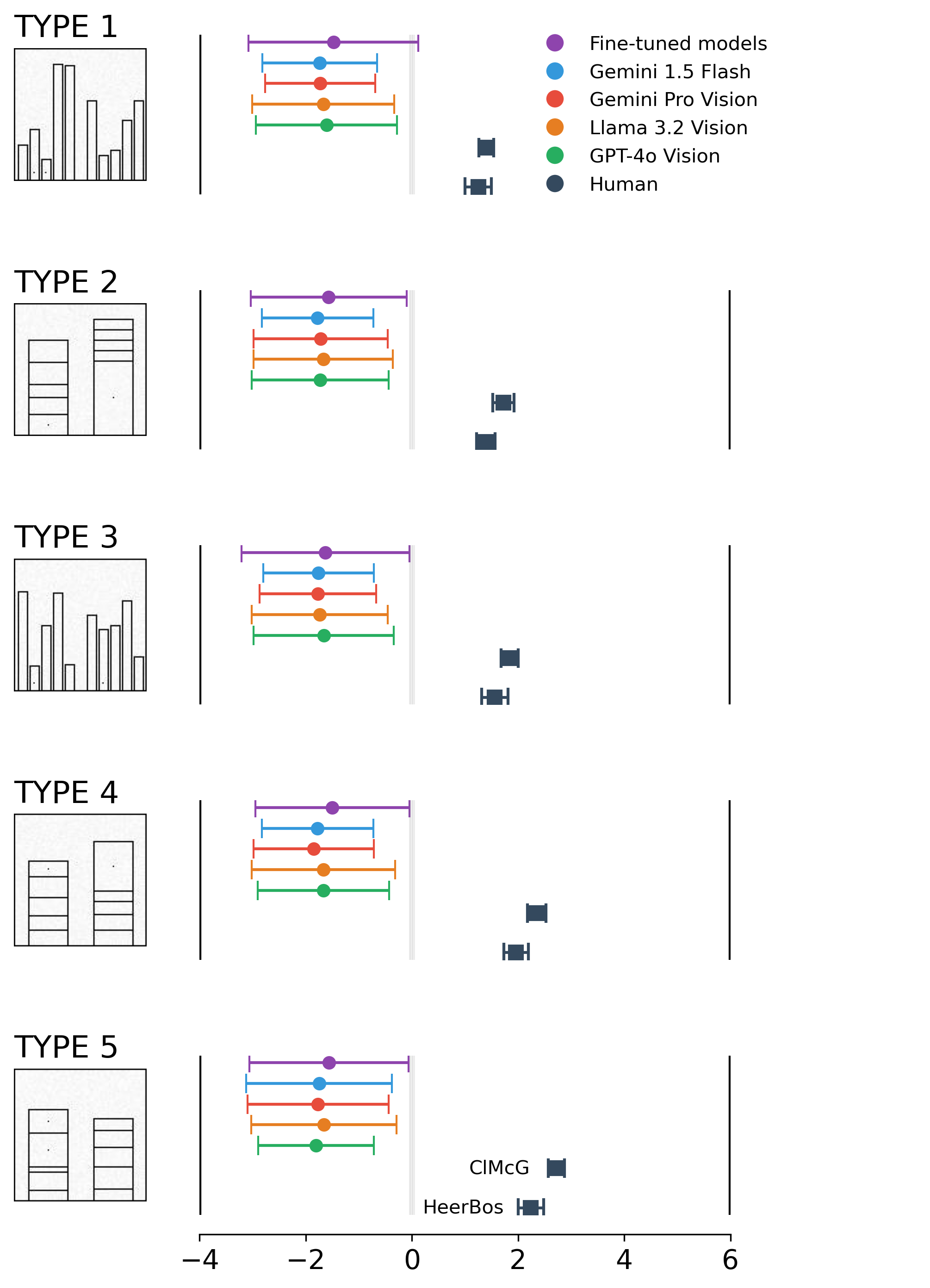}
    }
    \caption{\textbf{Computational results of the position-length experiment}. Left: Type 1–5 stimuli for divided and grouped bar charts (as per Cleveland and McGill). Right: MLAE and bootstrapped 95\% confidence
intervals of our networks.}
    \label{fig:enter-label}
\end{figure}

\subsection{Hypothesis and Results}

\textbf{H3.1: Pre-trained and fine-tuned MLLMs work well and outperform human perception for all five types of tasks.}

Our \textit{fine-tuned models} achieved ranges from (\textbf{MLAE: -1.63}, SD  =  0.74, MAE: 0.26) to (\textbf{MLAE: -1.48}, SD = 0.82, MAE: 0.29). Our \textit{pretrained models} recorded the lowest error with ranges of (\textbf{MLAE: -1.85}, SD =  0.53, MAE: 0.17) to (\textbf{MLAE: -1.61}, SD = 0.82, MAE: 0.29). In comparison to human baseline data, both fine-tuned and pretrained models produced the lowest errors, with \textbf{Human Baseline 1} \cite{Cleveland_84}: (MLAE: 1.4 to 2.72, SD from 0.14 to 0.175) and \textbf{Human Baseline 2} \cite{heer2010crowdsourcing}: (MLAE: 1.25 to 2.24, SD from  0.175 to 0.25). Also, Figure 4 displays that the MLAE of all MLLMs performs below zero, suggesting that \textcolor{blue} {\textbf{its performance estimates are mostly precise for five tasks}}. From the MLAE results above, we recognized that zero-shot prompting was effective for all pretrained models. The zero-shot prompts worked particularly well in these charts for all pretrained models. They specify the goal (e.g., compare bar heights), give detailed instructions on how to output answers (e.g., a scale from 0 to 1), and eliminate unnecessary complexity by using two words (no explanation). These prompts guide the models to concentrate solely on the visual comparison of bar and pie charts to produce single numeric outputs.

It also indicates that the general knowledge of pre-trained models enables them to recognize these tasks' structure. The lower average value of MLAE errors for this experiment might prove the pretrained models understand the concept of bar or pie charts and identify the relationship between different bars or pies  based on visual presentation. Also, the pretrained model might use its pretrained mathematical and logical reasoning to compute height and compare ratios. From this result, we \textbf{accept H3.1.}

\noindent \textbf{H3.2: Our fine-tuned models surpass pretrained models for all tasks.} 

\indent It is apparent from Figure 4 that all pretrained models consistently recorded lower MLAE scores than our fine-tuned models. \CHANGED {Across tasks, we evaluate the average regression performance of our networks and report statistically significant differences (\textbf{F = 36.66,} $p < 0.01\). Tukey’s HSD post-hoc test reveals that our fine-tuned models perform significantly worse than Gemini 1.5 Flash, Gemini Pro Vision,  Llama 3.2-Vision, and GPT-4o ($p < 0.01$). The mean differences range from \textbf{-0.13 to -0.22}, confirming that our fine-tuned models has the highest MLAE errors compared to all pretrained models. Since the differences are statistically significant across models},  we conclude that \textbf{our fine-tuned models underperform consistently and \textbf{thus we do not accept H3.2.}}

\section{Experiment 4}

\begin{figure}
    \hspace{1cm}
    \centering
    \scalebox{1.2}{%
        \includegraphics[width=0.8\linewidth]{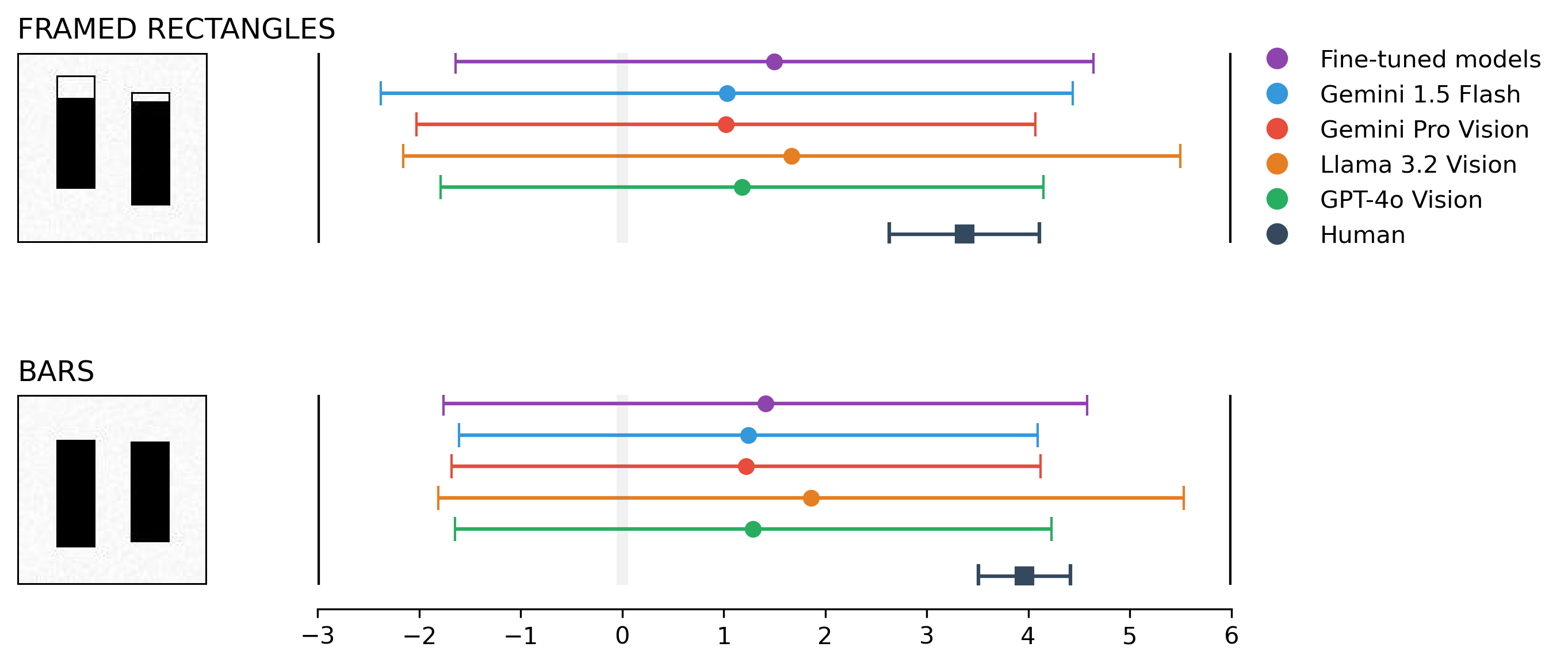}
    }
    \caption{\textbf{Computational results of the bars-and-framed-rectangles experiment}. Left: Stimuli of two bars for length judgment (bottom) following Cleveland and McGill’s setting. Perceiving which bar is longer is significantly easier for humans when a frame is added (top).}
    \label{fig:enter-label}
\end{figure}

Visual cues are essential in graphical elements as they integrate into real-world variables. Cleveland and McGill designed an experiment using bars and framed rectangles to study how humans perceive the length and position of non-aligned scales \cite {Cleveland_84}. Figure 5 highlights both variations on the left. It is difficult to estimate which bar is larger (bottom). Once the frame is added to the maximum length, the task transforms bar length into position judgment along aligned scales, making it easier to interpret. Cleveland and McGill theorized that judging the white space in the frame could resemble a length judgment rather than a position judgment. Given this, they relate the tasks to Weber's Law: the perceivable difference within a distribution is proportional to its initial size \cite {householder1940weberlaws}. In this experiment, Weber's Law implies that humans can find it easier to measure the differences in the whole space (framed scale) as its initial size is small. In contrast, estimating small changes in the length of black bars is harder. The Just Noticeable Difference (JND) is higher when the initial stimulus is smaller. We generated visualization charts aligned with these unique ground truth values for each framed and unframed task. 

\subsection{\textbf{Hypothesis and Results}}

\textbf{H4.1:}\textbf{ Fine-tuned and pretrained models surpass human perception for framed and unframed tasks.}

For \textit{framed tasks}, our fine-tuned model recorded average ranges of (\textbf{MLAE: 1.50}, SD = 1.60, MAE: 4.02). Our pretrained models have errors with average ranges of (\textbf{MLAE: 1.03}, SD = 1.74, MAE: 3.04) for Gemini 1.5 Flash, (\textbf{MLAE: 1.02, SD} = 1.56, MAE: 2.80) for Gemini 1.0 Pro Vision, (\textbf{MLAE: 1.18}, SD = 1.51, MAE: 3.09) for GPT-4o, and (\textbf{MLAE: 1.67, SD} = 1.95, MAE: 6.98) for Llama 3.2-Vision. For \textit{unframed tasks}, our fine-tuned models recorded average ranges of (\textbf{MLAE: 1.41}, SD = 1.62, MAE: 3.80). Our pretrained models have errors with average ranges of (\textbf{MLAE: 1.24}, SD = 1.46, MAE: 3.07) for Gemini 1.5 Flash, (\textbf{MLAE: 1.22}, SD = 1.48, MAE: 3.08) for Gemini 1.0 Pro Vision, (\textbf{MLAE: 1.29}, SD = 1.50, MAE: 3.30) for GPT-4o, and (\textbf{MLAE: 1.86}, SD = 1.87, MAE: 7.97) for Llama 3.2-Vision.

Our fine-tuned and pretrained models produced lower errors and outperformed human perceptions with \textbf{Human Baseline} (\textbf{MLAE:  3.371}, SD =  0.741) for framed tasks and \textbf{Human Baseline (MLAE: 3.961}, SD =  0.454) for unframed and framed tasks. In addition to this comparison, zero-shot prompting works effectively for framed and unframed tasks for pretrained models, especially Gemini 1.0 Pro Vision, as they have the lowest average MLAE errors, indicating those models generalize well without requiring any prior training. 

\textbf{H4.2: }\textbf{Our fine-tuned models work better than our pretrained models for two tasks. }

\indent Figure 5 highlights all pretrained models that produce lower average MLAE errors than our fine-tuned models for two tasks. However, an exception is shown here with the Llama 3.2-Vision model, \textcolor {blue} {\textbf{one of the pretrained models, which has underperformed our fine-tuned models}}. \CHANGED {Across tasks, we evaluate the average regression performance of our networks and report statistically significant differences (\textbf{F = 32.50,} \textbf{$p < 0.01$)}. Tukey’s HSD post-hoc test reveals that \textbf{our fine-tuned models performs significantly worse than Gemini 1.5 Flash, Gemini 1.0 Pro Vision, and GPT-4o ($p < =0.01$)}, with mean differences ranging from \textbf{-0.43 to -0.51}. However, its performance does not significantly differ from Llama 3.2-Vision ($p > 0.01$), indicating similar error levels between the two models. Since the differences are statistically significant across most models, we found that our fine-tuned models consistently underperforms except when compared to Llama 3.2-Vision.} Therefore, we \textbf{partially accept H4.2}. 

\begin{figure}
    \centering
    \hspace{1cm}
    \
    \scalebox{1.4}{%
        \includegraphics[width=0.8\linewidth]{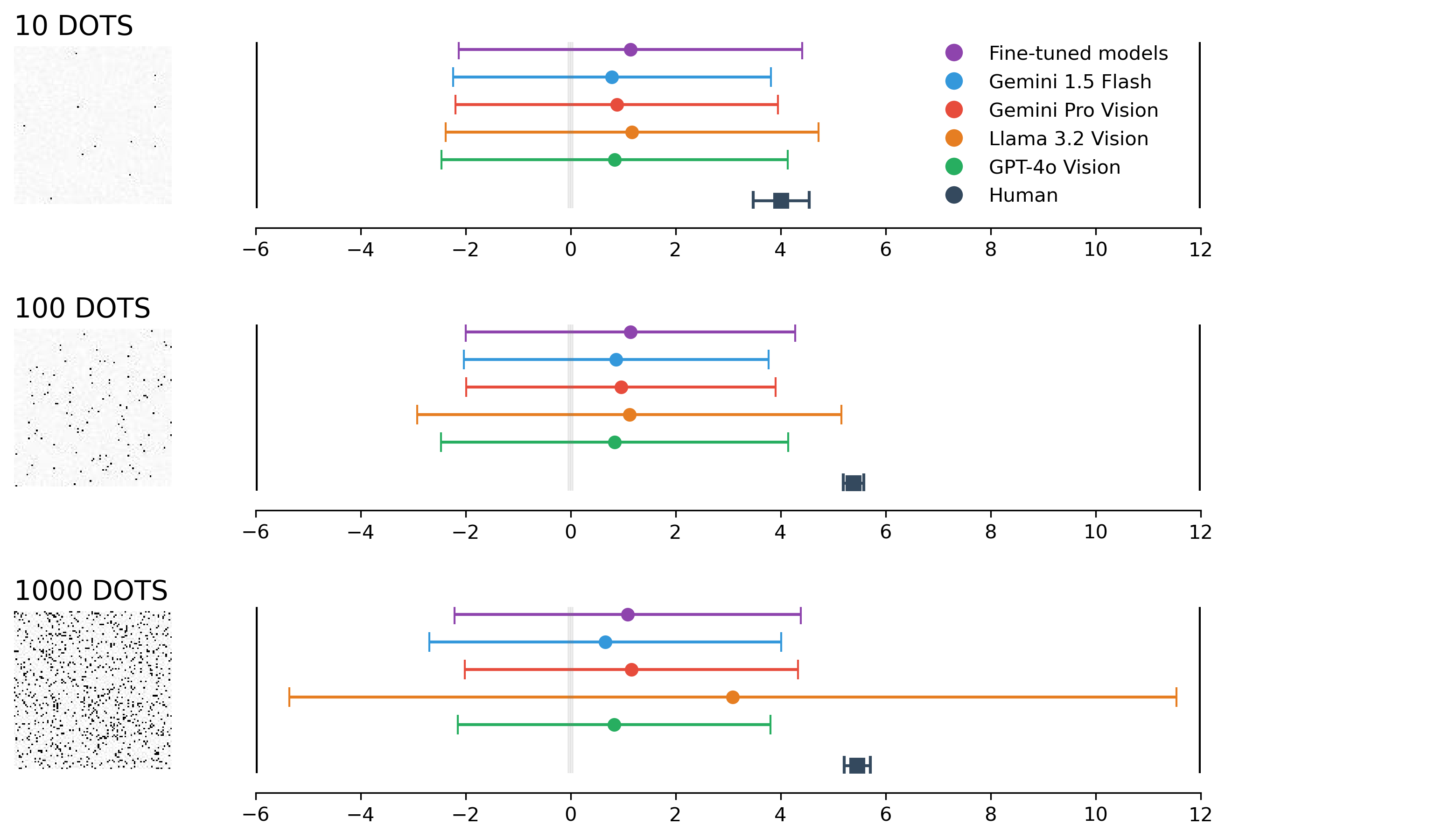}}
   
    \caption{\textbf{Computational results of the point cloud experiment.} Left: We create 2D point clouds with 10, 100, and 1000 initial dots. Then, we add up to 10 new dots. For humans, it is possible to estimate how many dots are added if there are initially 10 points, but it is impossible to see how many dots are added when starting with 1000 dots.}
    \label{fig:enter-label}
\end{figure}

\section {Experiment 5}

\subsection{\textbf{Hypothesis and Results}}

We generated a 2D point cloud simulation inspired by Weber’s law, where networks predict the number of dots (up to 10) added to an initial set of 10, 100, or 1000 dots. While humans can roughly estimate the number of added dots for 10 initial dots, it becomes harder for them to give answers and prone to random guessing with 100 or 1,000 initial dots. 

\textbf{H5.1: Our fine-tuned models perform relatively better than human perception and all pretrained models.}
\\\indent Our fine-tuned models performed across three tasks (10 dots, 100 dots, 1000 dots) with the following metrics:\textbf{10 dots: MLAE = 1.14}, SD = 1.67, MAE = 3.23; \textbf{100 dots: MLAE = 1.14}, SD = 1.60, MAE = 3.17; \textbf{1000 dots: MLAE = 1.09}, SD = 1.69, MAE = 3.12. These results significantly surpass the\textbf{ Human Baseline:} (\textbf{MLAE = 4.0149}, SD = 0.5338), (\textbf{MLAE = 5.3891}, SD = 0.1945), (\textbf{MLAE = 5.4612}, SD = 0.2509) for the respective tasks. In \textit{10 dots}, our pretrained models exceeded the range of an \textbf{MLAE from 0.79 to 0.88}, compared to our fine-tuned models at an \textbf{MLAE of 1.14}. Our fine-tuned models only outperformed Llama 3.2-Vision at an MLAE of 1.17. In \textit{100 dots}, our fine-tuned models with an \textbf{MLAE = 1.14}, SD = 1.60, MAE = 3.17 underperformed all pretrained models, with the range of average MLAE from 0.84 to 1.12. In \textit{1000 dots}, our fine-tuned models' performance with an \textbf{MLAE of 1.09}, SD = 1.69, MAE = 3.12 outperformed Llama 3.2-Vision, which recorded \textbf{MLAE = 3.09}, SD = 4.31, MAE = 503.27. 

\CHANGED {Across tasks, we evaluate the average regression performance of our networks and report statistically significant differences (\textbf{F = 37.44, $p < 0.01$}). Tukey’s HSD post-hoc test highlights that our fine-tuned models perform significantly worse than Gemini 1.5 Flash and GPT-4o (\textbf{$p < 0.01$}), with mean differences of \textbf{-0.35 and -0.29}, respectively. However, its performance does not significantly differ from Gemini 1.0 Pro Vision (\textbf{$p > 0.01$}), indicating similar error levels between the two models. Additionally, our fine-tuned models outperform Llama 3.2-Vision (\textbf{$p < 0.01$}) with a mean difference of \textbf{0.67}. Since the differences are statistically significant across most models, we found that our fine-tuned models generally underperform, except when compared to Gemini 1.0 Pro Vision and Llama 3.2-Vision.} Therefore, we \textbf{partially accept H5.1}.

\section{Conclusion}

Our paper reports the findings of evaluating the performance of MLLMs on graphical perception tasks in zero-shot prompt settings. This study measures pretrained and fine-tuned models' abilities across five experiments, focusing on elementary perceptual tasks, position length, position angle, position non-aligned scale, and point cloud. Overall, our pretrained models outperform fine-tuned models in most cases; all MLLMs can evaluate visualizations more precisely using zero-shot prompting on curvature tasks and all tasks from E2 to E5. 

Although our fine-tuned models mostly underperformed compared to pretrained models, the differences are minimal, and they perform well in volume and area tasks. Also, the post-hoc from our E2, highlights fine-tuned models that outperforms GPT-4o across bar and pie charts. This would be a strong opportunity for future research to fine-tune these models and enhance their performance. We trained our fine-tuned models with 94.4M parameters over five epochs using Parameter-Efficient Fine-Tuning (PEFT) and a LoRA configuration. Despite using fewer parameters, they performed closely to pretrained models, with only a small performance gap. These findings lay the groundwork for further enhancing fine-tuned models by scaling parameters and leveraging more diverse datasets. Since our MLLMs performed well in specific visual tasks, we wondered, “\textcolor {blue} {\textbf{Why do the MLLMs have wider MLAE error bars?}}”. This interesting point is explored in our supplement material. 

\CHANGED{Our findings also could be used for automating those zero-shot prompts to provide quicker insights for various real-world applications: business intelligence, scientific research, autonomous driving, robotics, etc. To achieve this purpose, our future studies will include more complex visualizations, diverse experiments, a wider range of MLLMs, color saturation and edge detection, and chain-of-thought reasoning to enhance MLLMs' ability to regress multiple values in visualizations.}




\bibliographystyle{abbrv-doi}
\bibliography{template.bib}

\begin{thebibliography}{99}

\bibitem{han2023chartllamamultimodalllmchart}
Y.~Han, C.~Zhang, X.~Chen, X.~Yang, Z.~Wang, G.~Yu, B.~Fu, and H.~Zhang,
  ``ChartLlama: A Multimodal LLM for Chart Understanding and Generation,''
  \emph{arXiv preprint arXiv:2311.16483}, 2023.

\bibitem{haehn2018evaluating}
D.~Haehn, J.~Tompkin, and H.~Pfister, ``Evaluating ‘graphical perception’ with
  CNNs,'' \emph{IEEE Transactions on Visualization and Computer Graphics},
  vol.~25, no.~1, pp. 641--650, 2018.

\bibitem{Cleveland_84}
W.~S. Cleveland and R.~McGill, ``Graphical perception: Theory,
  experimentation, and application to the development of graphical methods,''
  \emph{Journal of the American Statistical Association}, vol.~79, no.~387, pp.
  531--554, 1984.

\bibitem{cleveland1985graphical}
W.~S. Cleveland and R.~McGill, ``Graphical perception and graphical methods for
  analyzing scientific data,'' \emph{Science}, vol. 229, no. 4716, pp.
  828--833, 1985.

\bibitem{harrison2014ranking}
L.~Harrison, F.~Yang, S.~Franconeri, and R.~Chang, ``Ranking visualizations of
  correlation using Weber's law,'' \emph{IEEE Transactions on Visualization and
  Computer Graphics}, vol.~20, no.~12, pp. 1943--1952, 2014.

\bibitem{heer2010crowdsourcing}
J.~Heer and M.~Bostock, ``Crowdsourcing graphical perception: Using Mechanical
  Turk to assess visualization design,'' in \emph{Proceedings of the SIGCHI
  Conference on Human Factors in Computing Systems}, 2010, pp. 203--212.

\bibitem{Talbot2014FourEO}
J.~Talbot, V.~Setlur, and A.~Anand, ``Four experiments on the perception of bar
  charts,'' \emph{IEEE Transactions on Visualization and Computer Graphics},
  vol.~20, pp. 2152--2160, 2014.

\bibitem{zeng2024advancingmultimodallargelanguage}
X.~Zeng, H.~Lin, Y.~Ye, and W.~Zeng, ``Advancing multimodal large language
  models in chart question answering with visualization-referenced instruction
  tuning,'' \emph{arXiv preprint arXiv:2407.20174}, 2024.

\bibitem{agrawal2016vqavisualquestionanswering}
A.~Agrawal, J.~Lu, S.~Antol, M.~Mitchell, C.~L. Zitnick, D.~Batra, and
  D.~Parikh, ``VQA: Visual question answering,'' \emph{arXiv preprint
  arXiv:1505.00468}, 2016.

\bibitem{liu2024mmc}
F.~Liu, X.~Wang, W.~Yao, J.~Chen, K.~Song, S.~Cho, Y.~Yacoob, and D.~Yu,
  ``MMC: Advancing multimodal chart understanding with large-scale instruction
  tuning,'' in \emph{Proceedings of NAACL 2024}, pp. 1287--1310, 2024.

\bibitem{meng2024chartassisstantuniversalchartmultimodal}
F.~Meng, W.~Shao, Q.~Lu, P.~Gao, K.~Zhang, Y.~Qiao, and P.~Luo,
  ``ChartAssisstant: A universal chart multimodal language model via
  chart-to-table pre-training and multitask instruction tuning,'' \emph{arXiv
  preprint arXiv:2401.02384}, 2024.

\bibitem{du2023makesgoodvisualinstructions}
Y.~Du, H.~Guo, K.~Zhou, W.~X. Zhao, J.~Wang, C.~Wang, M.~Cai, R.~Song, and
  J.~R. Wen, ``What makes for good visual instructions? Synthesizing complex
  visual reasoning instructions for visual instruction tuning,'' \emph{arXiv
  preprint arXiv:2311.01487}, 2023.

\bibitem{han2024multimodallargelanguagemodels}
S.~C. Han, F.~Cao, J.~Poon, and R.~Navigli, ``Multimodal large language models
  and tunings: Vision, language, sensors, audio, and beyond,'' \emph{arXiv
  preprint arXiv:2410.05608}, 2024.

\bibitem{lv2024kosmos25multimodalliteratemodel}
T.~Lv \emph{et~al.}, ``KOSMOS-2.5: A multimodal literate model,'' \emph{arXiv
  preprint arXiv:2309.11419}, 2024.

\bibitem{lee2024multimodalreasoningmultimodalknowledge}
J.~Lee, Y.~Wang, J.~Li, and M.~Zhang, ``Multimodal reasoning with multimodal
  knowledge graph,'' \emph{arXiv preprint arXiv:2406.02030}, 2024.

\bibitem{wang2023visionllmlargelanguagemodel}
W.~Wang \emph{et~al.}, ``VisionLLM: Large language model is also an open-ended
  decoder for vision-centric tasks,'' \emph{arXiv preprint arXiv:2305.11175},
  2023.

\bibitem{liu2023visualinstructiontuning}
H.~Liu, C.~Li, Q.~Wu, and Y.~J. Lee, ``Visual instruction tuning,'' \emph{arXiv
  preprint arXiv:2304.08485}, 2023.

\bibitem{wang2024pictureworththousandwords}
J.~Wang \emph{et~al.}, ``Is a picture worth a thousand words? Delving into
  spatial reasoning for vision language models,'' \emph{arXiv preprint
  arXiv:2406.14852}, 2024.

\bibitem{xie2023agicomputervisionlessons}
L.~Xie \emph{et~al.}, ``Towards AGI in computer vision: Lessons learned from
  GPT and large language models,'' \emph{arXiv preprint arXiv:2306.08641},
  2023.

\bibitem{luo2024mmevolempoweringmultimodallarge}
R.~Luo \emph{et~al.}, ``MMEvol: Empowering multimodal large language models
  with evol-instruct,'' \emph{arXiv preprint arXiv:2409.05840}, 2024.

\bibitem{kafle2018dvqaunderstandingdatavisualizations}
K.~Kafle, B.~Price, S.~Cohen, and C.~Kanan, ``DVQA: Understanding data
  visualizations via question answering,'' \emph{arXiv preprint
  arXiv:1801.08163}, 2018.

\bibitem{guo2024understandinggraphicalperceptiondata}
G.~Guo, J.~J. Kang, R.~S. Shah, H.~Pfister, and S.~Varma, ``Understanding
  graphical perception in data visualization through zero-shot prompting of
  vision-language models,'' \emph{arXiv preprint arXiv:2411.00257}, 2024.

\bibitem{zhang2024exploringperceptuallimitationmultimodal}
J.~Zhang, J.~Hu, M.~Khayatkhoei, F.~Ilievski, and M.~Sun, ``Exploring perceptual
  limitation of multimodal large language models,'' \emph{arXiv preprint
  arXiv:2402.07384}, 2024.

\bibitem{bai2023qwenvlversatilevisionlanguagemodel}
J.~Bai \emph{et~al.}, ``Qwen-VL: A versatile vision-language model for
  understanding, localization, text reading, and beyond,'' \emph{arXiv
  preprint arXiv:2308.12966}, 2023.

\bibitem{wu2023vguidedvisualsearch}
P.~Wu and S.~Xie, ``V*: Guided visual search as a core mechanism in multimodal
  LLMs,'' \emph{arXiv preprint arXiv:2312.14135}, 2023.

\bibitem{kahou2018figureqaannotatedfiguredataset}
S.~E. Kahou \emph{et~al.}, ``FigureQA: An annotated figure dataset for visual
  reasoning,'' \emph{arXiv preprint arXiv:1710.07300}, 2018.

\bibitem{wang2024m2ptmultimodalprompttuning}
T.~Wang \emph{et~al.}, ``M$^2$PT: Multimodal prompt tuning for zero-shot
  instruction learning,'' \emph{arXiv preprint arXiv:2409.15657}, 2024.

\bibitem{householder1940weberlaws}
A.~S. Householder and G.~Young, ``Weber laws, the weber law, and psychophysical
  analysis,'' \emph{Psychometrika}, vol.~5, no.~3, pp. 183--193, 1940.

\end{thebibliography}

\end{document}